\documentclass{article}

\PassOptionsToPackage{numbers, compress}{natbib}

\usepackage[final]{neurips_2026}
\usepackage{wrapfig}
\usepackage{caption}

\usepackage[utf8]{inputenc} 
\usepackage[T1]{fontenc}    
\usepackage{hyperref}       
\usepackage{url}            
\usepackage{booktabs}       
\usepackage{amsfonts}       
\usepackage{nicefrac}       
\usepackage{microtype}      
\usepackage{xcolor}         
\usepackage{todonotes}
\title{An Empirical Study on Learning Rooftop Solar Potential from Satellite Imagery Directly Without Explicit Geometry}
\workshoptitle{2nd Workshop on Advances in Representation Learning for Earth Observation}

\newcommand{\cutsec}{\vspace{-1.7mm}}
\author{%
  Joshua Lee \\
  Skyline High School\\
  Ann Arbor, MI 48103 \\
  \texttt{joshua.ks.lee09@gmail.com} \\
}

\begin{document}

\maketitle

\begin{abstract}
  Rooftop solar potential estimation often relies on LiDAR, digital surface models (DSMs), or pipelines that explicitly reconstruct roof geometry before estimating solar potential. We propose Direct Rooftop Solar Prediction (DRSP), an end-to-end formulation that maps satellite imagery and coarse regional sunlight information directly to application-level solar outputs. DRSP neither requires explicit geometric representations as input nor predicts roof orientation, slope, height, DSMs, or other geometric quantities as intermediate outputs, and therefore requires no geometric intermediate supervision. To enable large-scale study of this formulation, we develop a programmatic data-curation pipeline that converts per-panel information from the Google Solar API into usable-rooftop masks and spatial solar-energy targets, yielding two geographically distinct datasets containing approximately 60K rooftops from the U.S. Midwest and 87K from Texas. We demonstrate the feasibility of DRSP and systematically characterize the effects of geographic training distribution, model depth, image resolution, regional sunlight conditioning, and heatmap regression objectives. Overall, our results suggest that useful rooftop-scale solar information can be learned directly from 2D imagery without explicit geometric intermediate representations, offering a simpler path toward large-scale solar assessment. Our code and dataset are available at \url{https://github.com/joshualee124/direct_rooftop_solar_prediction}.

  \vspace{-3mm}
\end{abstract}

\cutsec
\section{Introduction}
\cutsec

Mitigating climate change requires rapid decarbonization of the energy sector, and distributed rooftop photovoltaic (PV) generation can play an important role in expanding low-carbon electricity production. However, rooftop solar potential depends on factors such as usable roof area, orientation, slope, shading, and local irradiance. High-fidelity systems therefore often rely on LiDAR, digital surface models (DSMs), or explicit 3D roof reconstruction \cite{gassar2021review,melius2013estimating,google2026solar}, whose data and processing requirements can limit large-scale deployment.

Recent learning-based approaches reduce dependence on directly available 3D information by extracting rooftop regions before analytical PV modeling \cite{huang2019urban,zhong2021city,sun2022estimating}, or by predicting intermediate representations such as roof segments, orientation, slope, height, or DSMs for downstream solar estimation \cite{lee2019deeproof,li2023solarnet,boccalatte2025leveraging,batchu2024satellite,yan2023estimation}. Although the latter approaches need not consume geometry at inference time, they still require geometric intermediate prediction targets and corresponding supervision, which may depend on scarce annotations or specialized geospatial data \cite{krapf2022rid,boccalatte2025leveraging,batchu2024satellite}. An extended discussion of these related works is provided in Appendix~\ref{sec:extended_related_work}.

In this work, we propose \textbf{Direct Rooftop Solar Prediction (DRSP)}, which maps 2D RGB satellite imagery and coarse regional sunlight information directly to application-level solar outputs. Unlike pipelines that consume or reconstruct explicit rooftop geometry, DRSP neither requires LiDAR, DSMs, or other geometric representations as input nor predicts roof orientation, slope, height, or related quantities as intermediate outputs, and therefore requires no geometric intermediate supervision. We formulate DRSP to directly predict a usable-rooftop mask for panel placement and a spatial solar-energy heatmap, allowing geometry-related information needed for solar assessment to be learned implicitly from imagery.

To study DRSP at scale, we programmatically convert per-panel solar information from the Google Solar API \cite{google2026solar} into task-level supervision paired with satellite imagery, yielding two geographically distinct datasets: MIDWEST ($\sim$60K rooftops) and TEXAS ($\sim$87K). While the API provides supervision during dataset construction, DRSP itself is trained only on the resulting application-level targets rather than explicit geometric annotations.

Our key contributions are as follows: (1) we formulate DRSP, an end-to-end approach to rooftop solar prediction without explicit geometric inputs, intermediate representations, or supervision; (2) we develop a scalable data-curation pipeline yielding the geographically distinct MIDWEST and TEXAS datasets; (3) we systematically study DRSP across geographic distributions, model depths, image resolutions, regional sunlight conditioning, and heatmap regression objectives; and (4) we demonstrate the empirical viability of DRSP, showing that it approximates the ground truth usable-area and energy estimates within our dataset from RGB imagery alone and establishing a baseline for future work on direct rooftop solar prediction.




\cutsec
\section{Dataset Curation: MIDWEST and TEXAS}
\label{headings}
\cutsec

\begin{figure}
    \centering
    \includegraphics[width=1.05\linewidth]{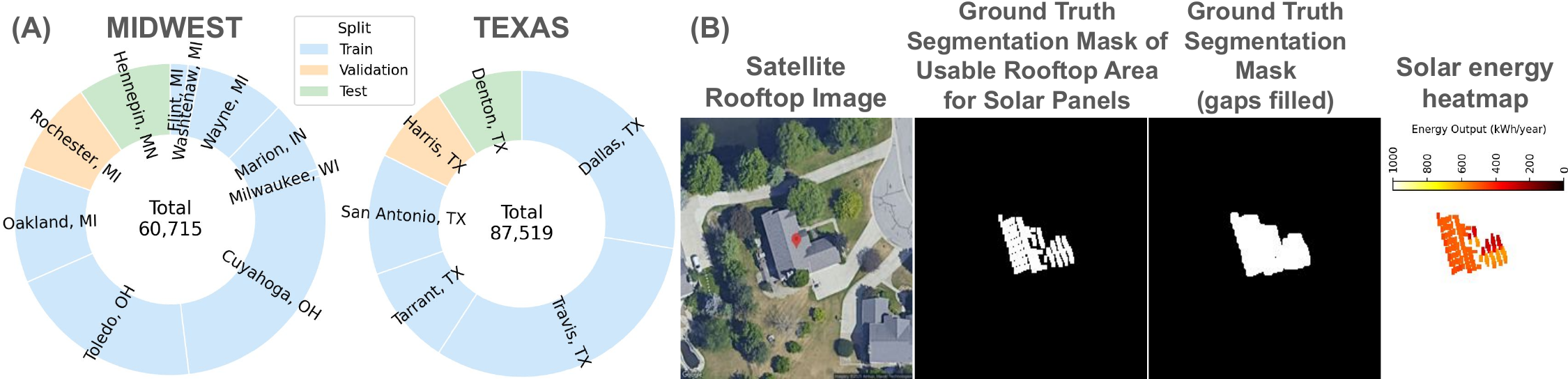}
    \vspace{-5mm}
    \caption{An illustration of our MIDWEST and TEXAS datasets. \textbf{(A)} Train, validation and test splits are divided by counties from within each dataset, and \textbf{(B)} A data instance example from MIDWEST.}
    \label{fig:data}
    \vspace{-1mm}
\end{figure}

We construct two rooftop solar datasets, \textbf{MIDWEST} and \textbf{TEXAS}, to study Direct Rooftop Solar Prediction (DRSP) across geographically distinct U.S. regions. For each rooftop, we pair a 640x640 satellite image with rooftop-level solar information obtained from the Google Solar API \cite{google2026solar}.

Supervision targets are derived programmatically from the API's \emph{per-panel} information, which provides the locations of recommended solar panels and their estimated annual energy production.  We rasterize the panel locations to construct (1) a gap-filled binary segmentation mask indicating rooftop regions suitable for panel placement and (2) a spatial energy heatmap in which panel-covered pixels are assigned the corresponding estimated energy generation. This enables dense supervision without manual annotation of roof orientation, slope, height, DSMs, or other explicit geometric intermediate representations. An example data instance is shown in Figure~\ref{fig:data}B. From now on, we refer to these API-derived references (more information in Appendix B) as the ground truth. 

MIDWEST contains approximately 60K rooftops from 10 counties, while TEXAS contains approximately 87K rooftops from 6 counties, as shown in Figure~\ref{fig:data}A. We use geographically disjoint training, validation, and test regions rather than random rooftop-level splits, enabling evaluation of both within-region performance and generalization across geographic distributions. Full curation details, target construction procedure, and per-county statistics are provided in Appendix~\ref{app:data}. 

\cutsec
\section{Methodology}
\cutsec

\begin{figure}
    \centering
    \includegraphics[width=1.05\linewidth]{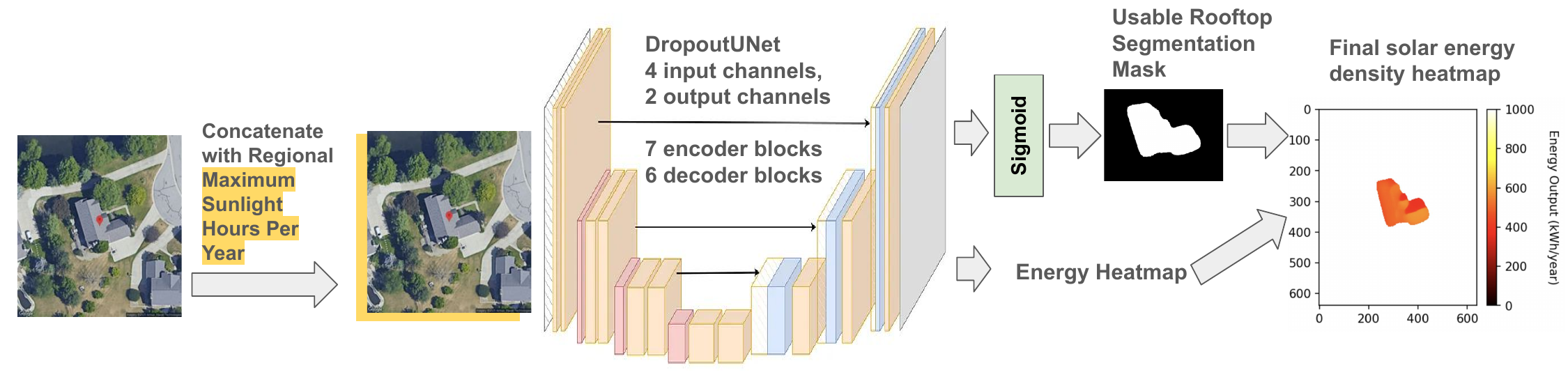}
    \vspace{-7.5mm}
    \caption{Illustration of our methodology. We first concatenate the maximum sunlight hour of the region to the input satellite image, then pass it through a DropoutUNet that outputs a usable rooftop segmentation mask and an energy heatmap, which are overlaid to produce the final heatmap.}
    \label{fig:method}
    \vspace{-1mm}
\end{figure}

We instantiate DRSP using a DropoutUNet-based~\cite{dropoutunet} dense prediction model, which we refer to as \textbf{DRSP-Net}. As illustrated in Figure~\ref{fig:method}, the model receives a satellite RGB image together with the corresponding regional maximum annual sunlight hours. The sunlight value is encoded as an additional constant-valued input channel, producing a four-channel input to the network.

DRSP-Net jointly predicts two spatial outputs: a usable-rooftop segmentation mask and a solar-energy heatmap. The segmentation branch is trained with a binary classification objective, while the energy branch is trained only over pixels for which solar-energy supervision is available using an L1 regression objective. The two objectives are optimized jointly. From these dense outputs, we additionally derive rooftop-level quantities such as maximum panel count and total annual energy generation using simple linear calibration factors estimated from the training set. Together, these outputs and derived quantities capture the key information needed for practical applications such as rooftop solar panel placement and energy-generation assessment.

Our goal is not to introduce a specialized architecture, but to use this formulation as a controlled testbed for studying DRSP. We therefore systematically vary model depth, image resolution, heatmap regression objective, and geographic data composition in Section~\ref{sec:experiments}. Architectural details, preprocessing, output construction, loss definitions, calibration factor details, and training hyperparameters are provided in Appendix~\ref{app:method}.

\cutsec
\section{Experiments}
\label{sec:experiments}
\cutsec

We first confirmed the feasibility of DRSP by training DRSP-Net on training splits of both MIDWEST and TEXAS, and we found that the model produces rooftop heatmaps that are generally consistent with the ground truth. Qualitative examples (Figure~\ref{fig:examples}, and Figure~\ref{fig:qualapp} in the Appendix) show that DRSP-Net learns roof-aligned spatial patterns from 2D imagery despite receiving no explicit geometric intermediate supervision. In several examples, predicted energy also decreases near visually apparent occlusions such as trees.

\begin{figure}
    \centering
    \includegraphics[width=1.0\linewidth]{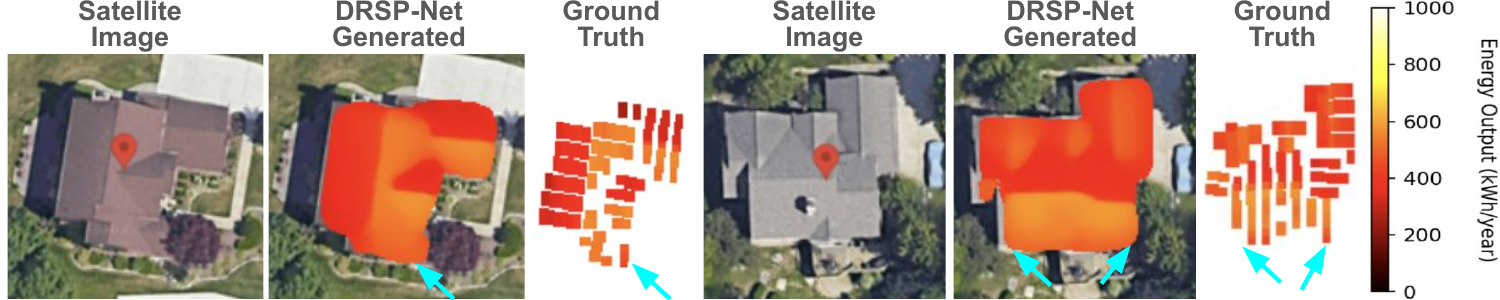}
    \vspace{-5.5mm}
    \caption{Qualitative examples of DRSP-Net generation. Our model learns roof-aligned spatial patterns without explicit geometric supervision, and is even able to predict lower generation for rooftop regions near trees (\textcolor{cyan}{cyan arrows}).}
    \label{fig:examples}
    \vspace{-3mm}
\end{figure}

\cutsec
\subsection{Impact of geographical distribution of training data}
\cutsec

A key question for DRSP is how geographic differences in training data affect generalization, since rooftop appearance and solar potential can vary substantially across regions. We compare DRSP-Net trained on MIDWEST only, TEXAS only, and both datasets jointly, with results shown in Table~\ref{tab:d1}. For the two rooftop-level numerical predictions, calibration factors are always estimated using training data from the corresponding evaluation region, independent of the DRSP-Net training data.

Joint training on both regions achieves the best performance on all metrics across both test sets except energy heatmap MAE for MIDWEST. In particular, it improves rooftop segmentation and panel-count prediction relative to region-specific training, suggesting that geographically diverse data can improve learning of region-invariant visual structure. For total energy generation, the jointly trained model also performs best after local calibration, while pixel-wise energy heatmap MAE is quite close between BOTH and the model trained only on the corresponding region, indicating that a shared visual model can still make accurate region-specific predictions with or without local calibrations.  

MIDWEST-only and TEXAS-only models degrade substantially when evaluated on the other region. These results indicate that spatial energy prediction is more sensitive to geographic distribution shift than rooftop geometry, and suggest that scaling DRSP across regions benefits from geographically diverse training data together with local environmental conditioning or calibration.

\cutsec
\subsection{Model architecture and objective empirical study}
\cutsec

We further study how network depth, input resolution, regional sunlight conditioning, and the heatmap regression objective affect DRSP performance on MIDWEST (Table~\ref{tab:abla}). Deeper U-Nets generally perform better, while reducing the input resolution substantially degrades both segmentation and downstream numerical prediction. Removing regional sunlight hours leaves segmentation performance nearly unchanged but worsens energy-related metrics, suggesting that coarse irradiance information provides complementary signal. For heatmap regression, L2 produces blurrier, lower-contrast heatmaps (Figure~\ref{fig:yourfigure}) and worse panel-count and total-energy predictions, so we use L1 by default. Finally, directly regressing panel count and total energy with a fine-tuned SatMAE~\cite{cong2022satmae} performs substantially worse than DRSP-Net, suggesting that learning spatial rooftop usability and energy distributions is beneficial for accurate whole-rooftop prediction.

\begin{table}[]
\small
\centering
\setlength{\tabcolsep}{2pt}
\begin{tabular}{@{}l|cccc|cccc@{}}
\toprule
 & \multicolumn{4}{c|}{MIDWEST Test Set}
 & \multicolumn{4}{c}{TEXAS Test Set} \\ \cline{2-9}

\begin{tabular}[c]{@{}l@{}}DRSP-Net\\Training\\ Data\end{tabular}
& \begin{tabular}[c]{@{}c@{}}Usable\\ Rooftop\\ Segmentation\\ MIOU $\uparrow$\end{tabular}
& \begin{tabular}[c]{@{}c@{}}Energy\\ Heatmap\\ Pixel-wise \\ MAE $\downarrow$\end{tabular}
& \begin{tabular}[c]{@{}c@{}}Max\\ Panel\\ Count\\ MAE $\downarrow$\end{tabular}
& \begin{tabular}[c]{@{}c@{}}Total\\ Energy\\ Generation\\ MAE $\downarrow$\end{tabular}
& \begin{tabular}[c]{@{}c@{}}Usable\\ Rooftop\\ Segmentation\\ MIOU $\uparrow$\end{tabular}
& \begin{tabular}[c]{@{}c@{}}Energy\\ Heatmap \\ Pixel-wise \\ MAE $\downarrow$\end{tabular}
& \begin{tabular}[c]{@{}c@{}}Max\\ Panel\\ Count\\ MAE $\downarrow$\end{tabular}
& \begin{tabular}[c]{@{}c@{}}Total\\ Energy\\ Generation\\ MAE $\downarrow$\end{tabular} \\
\hline

BOTH & \textbf{0.67} & 0.045 & \textbf{17.45} & \textbf{8410.46}
     & \textbf{0.77} & \textbf{0.034} & \textbf{18.73} & \textbf{10936.67} \\

MIDWEST only & 0.66 & \textbf{0.043} & 18.30 & 8962.27
             & 0.66 & 0.062 & 37.50 & 20939.09 \\

TEXAS only & 0.53 & 0.078 & 35.37 & 17447.87
           & 0.76 & 0.035 & 19.71 & 11147.66 \\

\bottomrule
\end{tabular}
\vspace{1mm}
\caption{Experiment results on training DRSP-Net with different training datasets.}
\label{tab:d1}
\vspace{-1mm}
\end{table}

\begin{figure*}[t]
    \centering

    \begin{minipage}[t]{0.65\textwidth}
\vspace{-6pt}
        \small
\centering
\setlength{\tabcolsep}{2pt}

\begin{tabular}{@{}l|ccccc@{}}
\toprule
MIDWEST-only
& \begin{tabular}[c]{@{}c@{}}Usable\\ Rooftop\\ Segmentation\\ MIOU $\uparrow$\end{tabular}
& \begin{tabular}[c]{@{}c@{}}Energy\\ Heatmap \\ Pixel-wise \\ MAE $\downarrow$\end{tabular}
& \begin{tabular}[c]{@{}c@{}}Max\\ Panel\\ Count\\ MAE $\downarrow$\end{tabular}
& \begin{tabular}[c]{@{}c@{}}Total\\ Energy\\ Generation\\ MAE $\downarrow$\end{tabular} \\
\hline

DRSP-Net (7-layer, L1)
&0.66 & \textbf{0.043} & \textbf{18.30} & \textbf{8962.27} \\

6-layer
&0.65 &0.049 &20.15 &9694.80 \\

5-layer
& 0.60 & 0.048 & 22.81 &10676.00 \\

4-layer
&0.50&0.052 &25.98 &11923.19 \\

No sunlight hours
&0.66 &0.050 & 18.87 & 9110.98 \\

224x224 resolution

&0.27 &0.063 &44.83 &20929.91 \\

L2 heatmap objective
&\textbf{0.68} & 0.047 &22.85 &11351.37 \\

Finetuned SatMAE~\cite{cong2022satmae}
& - & -  & 58.23 & 29423.77 \\

\bottomrule
\end{tabular}
\vspace{0mm}
\captionof{table}{Results on training DRSP-Net on MIDWEST only, compared with ablations in model design, layers, image resolution, and training objective, as well as alternative prediction methods.}
\label{tab:abla}
    \end{minipage}
    \hfill
    \begin{minipage}[t]{0.32\textwidth}
    \vspace{-1pt}
        \centering
        \includegraphics[width=1.1\linewidth]{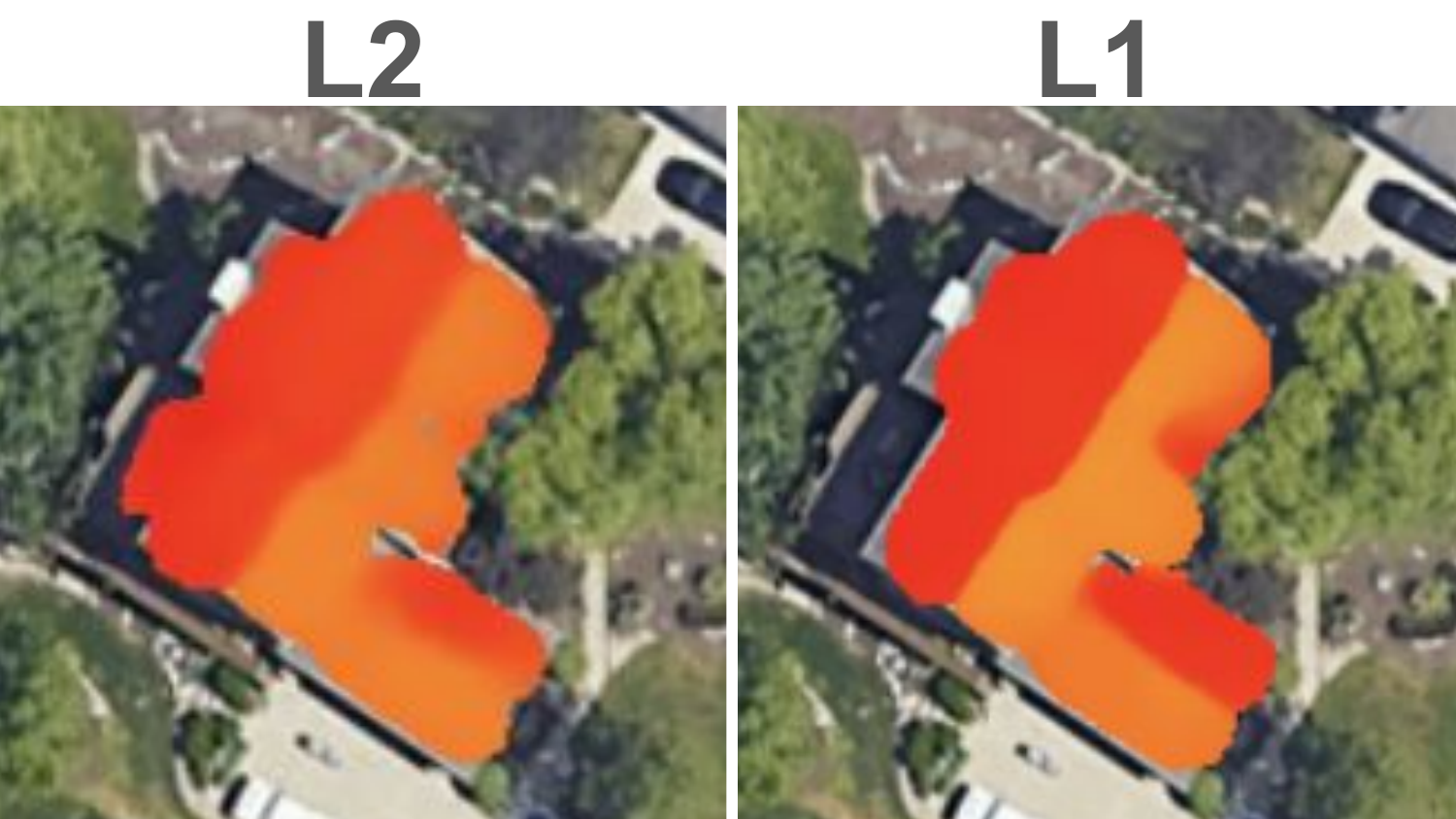}
        \captionof{figure}{Comparison of L1 objective with L2 objective. L2 objective's heatmaps are generally more blurry and have weaker contrast.}
        \label{fig:yourfigure}
    \end{minipage}

\vspace{-3mm}
\end{figure*}

\cutsec
\section{Conclusion, Limitations and Future Work}
\cutsec

We propose Direct Rooftop Solar Prediction (DRSP), an end-to-end formulation for estimating rooftop solar potential directly from satellite imagery and coarse regional sunlight information without explicit geometric inputs, intermediate representations, or supervision. We develop the programmatically generated MIDWEST and TEXAS datasets and demonstrate the feasibility of DRSP across geographically distinct regions, while characterizing key modeling and data choices that affect performance. Overall, our results suggest that application-relevant rooftop solar information can be learned directly from 2D imagery while leaving geometry-related information implicit, offering a simpler path toward large-scale solar assessment. Our current evaluation is limited to selected U.S. regions; however, the data-curation pipeline is in principle applicable across much broader geographic areas, motivating future work on nationwide scaling and stronger regional conditioning.

{\small
\bibliographystyle{unsrtnat} 
\bibliography{references}
}

\clearpage

\appendix

\section{Extended Related Work}
\label{sec:extended_related_work}

Rooftop solar potential assessment commonly combines remote sensing, geometric modeling, and PV simulation. Prior approaches broadly rely on explicit geometric representations, use learning to extract rooftop regions before analytical estimation, or predict geometric intermediate representations for downstream solar modeling. DRSP instead studies whether application-level solar outputs can be predicted directly from 2D imagery while leaving geometry-related information implicit.

\paragraph{Geometry-based and rooftop-extraction approaches.}
High-fidelity systems often use LiDAR, digital surface models (DSMs), photogrammetry, or reconstructed 3D building models to obtain roof area, slope, orientation, height, and shading information \cite{gassar2021review,melius2013estimating}. These representations enable detailed physical modeling but introduce additional data and processing requirements. Learning-based methods have reduced this dependence by extracting rooftop regions from imagery before applying analytical PV estimation. Huang et al.\ \cite{huang2019urban}, Zhong et al.\ \cite{zhong2021city}, and Sun et al.\ \cite{sun2022estimating} use deep segmentation or rooftop-classification models to identify suitable rooftop areas, followed by separate estimation procedures based on rooftop area, roof type, irradiance, or panel-layout assumptions. In contrast, DRSP directly predicts usable rooftop regions and spatial solar-energy outputs rather than using learned rooftop features as inputs to a separate analytical model.

\paragraph{Predicting explicit geometric intermediates.}
Other methods infer rooftop geometry from imagery but retain geometry as an explicit intermediate representation. DeepRoof \cite{lee2019deeproof} predicts planar roof segments and orientations before solar-potential estimation, while SolarNet \cite{li2023solarnet} predicts roof segmentation and geometric roof representations for downstream assessment. SolarMTNet \cite{boccalatte2025leveraging} jointly predicts roof orientation and slope, and Yan et al.\ \cite{yan2023estimation} predict rooftop regions and building heights to construct 3D building models. Satellite Sunroof \cite{batchu2024satellite} predicts high-resolution DSMs and roof segmentation from satellite imagery for large-scale solar mapping. Related recent work further models rooftop superstructures or combines RGB imagery with nDSM information \cite{li2024solarnetplus,yao2026solarfusion}. These approaches reduce the need for directly observed geometry at inference time, but still explicitly predict or consume geometric quantities such as orientation, slope, height, or DSMs.

\paragraph{Geometric supervision and DRSP.}
Explicit geometric prediction also requires suitable supervision. RID \cite{krapf2022rid}, for example, provides detailed annotations of roof segments and rooftop superstructures, while other methods derive orientation, slope, DSM, or segmentation labels from auxiliary geospatial data \cite{boccalatte2025leveraging,batchu2024satellite}. Our pipeline instead derives task-level supervision from per-panel recommendations and energy estimates from the Google Solar API \cite{google2026solar}, without separately constructing labels for roof orientation, slope, height, DSMs, or related quantities. This does not imply that the API itself operates without geometry; rather, explicit geometry is absent from the inputs, intermediate outputs, and supervision required by the learned DRSP model.

\paragraph{Positioning of DRSP.}
The key distinction is therefore not simply that DRSP uses RGB imagery. Several prior methods also infer solar-relevant information from imagery. Instead, DRSP differs in its prediction formulation: geometry-oriented methods explicitly reconstruct roof properties and then use them for solar estimation, whereas DRSP maps imagery and coarse regional sunlight information directly to application-level outputs. Explicit geometry offers greater physical interpretability and may support downstream analyses beyond solar prediction; DRSP instead studies whether the information needed for usable-area, spatial-energy, panel-count, and total-energy prediction can be learned implicitly without geometric intermediate supervision.

\paragraph{Relation to EO representation learning and foundation models.} 
A separate line of work pretrains general-purpose EO representations intended to transfer across many downstream tasks, rather than targeting a single application: SatMAE~\cite{cong2022satmae}, Prithvi~\cite{jakubik2023prithvi}, and Clay~\cite{clay2024foundation} pretrain broadly transferable geospatial representations, while Scale-MAE~\cite{reed2023scalemae} targets scale-invariant representations across varying ground sampling distances. These models encode geometric structure implicitly, but independently of any specific downstream task. Our SatMAE comparison (Section~4.2) suggests that for a fine-grained, geometry-dependent task such as rooftop-scale solar estimation, a representation learned end-to-end under dense task supervision captures spatial cues that a fine-tuned, general-purpose EO representation does not fully preserve.

\section{Data curation details}
\label{app:data}

For each county, the data curation process began by obtaining building coordinates from OpenAddress Batch Processing~\cite{openaddresses_batch_data_2026}. These initial coordinates were passed to the Google Solar API~\cite{google2026solar} to retrieve detailed solar potential data in JSON format. Because the initial coordinates from OpenAddress Batch Processing~\cite{openaddresses_batch_data_2026} were slightly inaccurate, we extracted the precise center coordinates directly from the resulting solar JSON files to ensure accurate alignment for image download, where the target rooftop is always centered. These refined coordinates were then used to retrieve high-resolution satellite imagery via the Google Maps API at a zoom level of 20. The images were downloaded with 640x640 resolution. 

We use dictionary indexing from the Google Solar API~\cite{google2026solar} to access raw solar data for each address, from which we were able to obtain the locations of optimal solar panels, the number of solar panels that can fit on the roof, and the energy output at each panel location. From this information, two primary ground truth outputs were generated: (1) a binary segmentation mask of rooftop area suitable for solar panel placement, and (2) a partial heatmap over solar energy generation potential. Using the precise latitude and longitude bounds extracted from the API (applying a latitude padding of approximately 0.00032 and a longitude padding of 0.00043), geographical coordinates for individual solar panels were converted into pixel-level coordinates within the 640×640 image space. Solar panels were represented as rectangles based on their specified orientation (portrait or landscape). There were random small gaps between the solar panels, and they were filled 
to provide a continuous spatial mask of the rooftop area suitable for panel placement. 
The energy heatmap was generated by mapping the yearly energy production (kWh/year) of each panel to the corresponding pixels in the mask. For pixels covered by multiple panels, energy values were averaged to reflect local density. Pixels not covered by any solar panel were assigned a value of -1, serving as an ignore index during model training to focus the loss function exclusively on rooftop areas. Our experiments have shown that, even when trained with heatmaps where annotations were available on some of the rooftop pixels, our model was able to generate accurate heatmaps covering all pixels of the rooftop.

To study the effect of geographical differences in training and model generalizability, we curated the MIDWEST dataset on 10 counties in the Midwest region of the United States, and TEXAS dataset on 6 counties in Texas, covering northern and southern regions of the contiguous United States. The detailed number of rooftops per county are listed in Table~\ref{tab:data}. The regional sunlight hours information, as well as the ground truth maximum panel count and total energy generation (in kWh/year) was also obtained from Google Solar API~\cite{google2026solar}.

Note: The Google Solar API derives its estimates from a geometry-based pipeline. It first builds a digital surface model (DSM) of each building from high-resolution aerial imagery, or from satellite imagery with machine-learned elevation estimates where aerial coverage is unavailable \cite{batchu2024satellite}. It then segments the roof into planar sections with estimated pitch and azimuth. Using this 3D model, the API simulates the sun's position across the year together with local weather data and shading from nearby structures and vegetation, producing an annual solar flux map over the roof surface. Panels are then placed on the roof segments, prioritizing high-flux areas while avoiding obstructions, and each panel's expected yearly energy production is computed from the flux it receives and a reference panel specification. None of the API's intermediate geometric products, including the DSM, roof segments, pitch and azimuth estimates, and flux maps, are used by our model as inputs, intermediate prediction targets, or supervision. We use only the final per-panel layouts and energy values, which we rasterize into our segmentation and heatmap targets. These targets still implicitly reflect roof geometry, orientation, and shading, but DRSP must learn this information from RGB imagery alone.

\section{Methodology details}
\label{app:method}


\subsection{Input preprocessing} 

Each satellite image (3-channel RGB image) with size 3x640x640 was converted to a Torch Tensor, and was normalized using ImageNet~\cite{deng2009imagenet} mean and std. Subsequently, a fourth channel was added to the image, in which each pixel value was set to the same value (the region's maximum annual sunlight hours, divided by 2000).

\subsection{Model Architecture}

Our main model is a DropoutUNet~\cite{dropoutunet}. In our model, a "convolution block" is composed of two stacked convolutional layers, each using a 3x3 kernel with stride 1 and padding 1 to maintain spatial dimensions. We applied batch normalization after each convolution to normalize feature distributions. We applied ReLU activations after batch normalization to introduce nonlinearity. Spatial dropout was applied after the ReLU activation to prevent overfitting (i.e., the model memorizing the training data).
Our model comprises seven encoder convolutional blocks and six decoder convolutional blocks. The input passes through the encoder blocks sequentially, with 2x2 max pooling applied between blocks to reduce feature map size. The seven encoder blocks progressively with 32, 64, 96, 128, 160, 192, and 224 channels, respectively. Next, six decoder blocks progressively reconstruct the feature maps' spatial resolution using transposed convolutions. Before each decoder block, a 2×2 transposed convolution with a stride of 2 doubles the spatial dimensions while reducing the number of feature channels, and the upsampled feature maps are concatenated with the corresponding encoder feature maps through skip connections.

\begin{table}
\centering
\begin{tabular}{llll}
\toprule
County/City & Dataset &  Data Split & Rooftop Counts \\ \hline
Flint, MI & MIDWEST & Train & 1129 \\
Washtenaw, MI & MIDWEST & Train & 792 \\
Wayne, MI & MIDWEST & Train & 5500 \\
Marion, IN & MIDWEST & Train & 4555 \\
Milwaukee, WI & MIDWEST & Train & 455 \\
Cuyahoga, OH & MIDWEST & Train & 16754 \\
Toledo, OH & MIDWEST & Train & 12370 \\
Oakland, MI & MIDWEST & Train & 7343 \\
Dallas, TX & TEXAS & Train & 24132 \\
Travis, TX & TEXAS & Train & 27598 \\
Tarrant, TX & TEXAS & Train & 9244 \\
San Antonio, TX & TEXAS & Train & 11142 \\
Rochester, MI & MIDWEST & Validation & 5900 \\
Harris, TX & TEXAS & Validation & 7372 \\
Hennepin, MN & MIDWEST & Test & 5917 \\
Denton, TX & TEXAS & Test & 8031 \\
\bottomrule
\end{tabular}
\caption{Dataset curation details: counties used, number of rooftops per county, and their data splits}
\label{tab:data}
\end{table}

\subsection{Output Processing and Training Objectives}

The DropoutUNet's 2-channel outputs predict, respectively, the areas of the roof suitable for solar panel placement and the solar energy density at each location. We pass the first channel output through a sigmoid layer. We have separate loss functions for the 2 outputs: we use a BCELoss between the first channel outputs (post-sigmoid) and the ground truth mask, and an L1 Loss between all pixels in the ground truth heatmap that are not -1 and the same pixels in the second channel output. Therefore, the first channel is trained to predict rooftop segmentation masks, and the second channel is trained to generate a heatmap of solar panel energy potential at each pixel. The overall training objective is a weighted sum of the two losses (1x segmentation loss + 10x heatmap loss).

\begin{figure}
    \centering
    \includegraphics[width=1.0\linewidth]{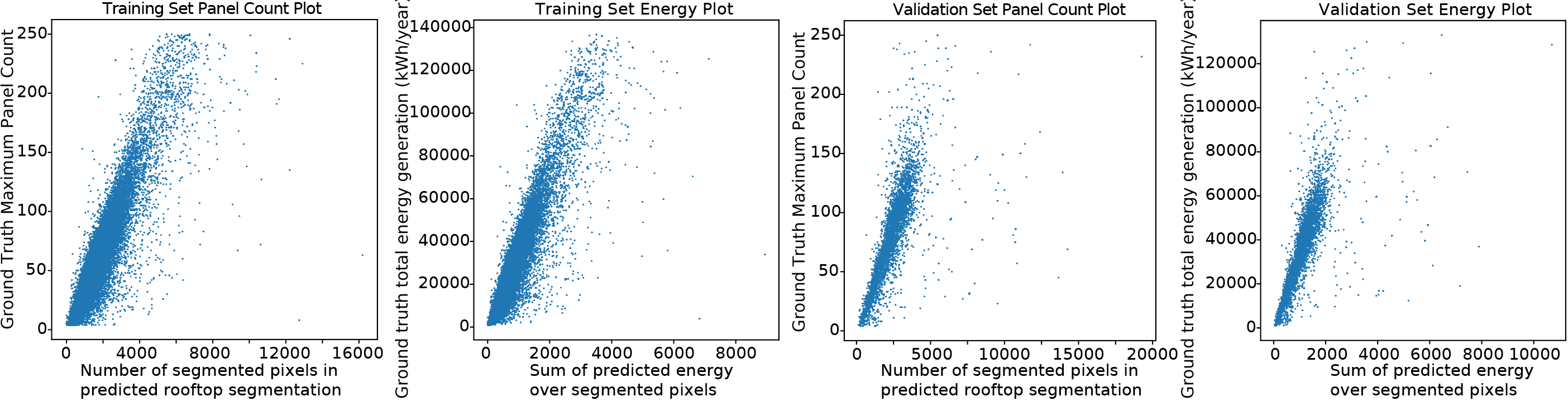}
    \caption{Scatter plots of maximum panel count prediction and total energy prediction on both the training split and the validation split. The "panel count" plots show the relationship between the number of segmented pixels in the predicted segmentation mask of the usable rooftop area for solar panel placement and the ground truth maximum number of panels recommended for the rooftop; and the "energy" plots show the relationship between the sum of energy in the predicted heatmap over the segmented pixels and the ground truth total energy generation over all recommended panels. The scatter plots clearly show strong linear correlations across both training and validation splits, thereby justifying our divisor-based approach to numerical prediction.}
    \label{fig:scatter}
\end{figure}

In addition, we use the model's outputs to predict maximum panel count and total energy generation. The binary segmentation masks were obtained by including all pixels higher than a threshold (0.5) in the post-sigmoid first-channel output, and the total number of segmented pixels was divided by a constant to predict panel count. We obtained the divisor as the mean ratio between the segmented pixel count and the ground-truth panel count on the training set. Similarly, we sum the energy density prediction values (i.e., the second channel output) over each segmented pixel and divide by a constant to predict total energy generation. The divisor was again obtained by the mean ratio between the sum and ground-truth total energy generation on the training set. We exclude all outlier rooftops with a ground-truth maximum panel count greater than 250 across all data splits when determining divisors and evaluating metrics. We use this divisor-based method because we found that these values are generally directly proportional (see Figure~\ref{fig:scatter}). We have also tried other more complex approaches like fine-tuning a small CNN to predict these values based on segmentation map and heatmap, but found that they do not predict more accurately than the simple divisor-based method.

\subsection{Training details and evaluation metrics}

The model was optimized using the Adam~\cite{kingma2014adam} optimizer, chosen for its fast convergence when training deep neural networks. The model was trained with a batch size of 96 and a learning rate of 1e-4. The Adam optimizer was used with a weight decay of 1e-4 to provide L2 regularization and reduce overfitting by discouraging large parameter values. Several rounds of hyperparameter tuning (varying the learning rate and batch size to improve performance) were conducted to identify the optimal hyperparameter values listed above, based on validation set performance. The dropout rate was 0.2. Models were trained for 200 epochs, and then its performance was evaluated on validation data. Models were trained for 200 epochs and evaluated on the validation set during training. The best-performing checkpoint on the validation data was used to report performance over the test set. We report the following metrics: mIoU (Mean Intersection Over Union) was used for segmentation performance; pixel-wise Mean Absolute Error (MAE) was used for heatmap generation quality (the unit for this metric is the same as the raw pixel annotation values in the processed ground truth heatmaps, i.e. the power generation capacity of the entire solar panel that this pixel belongs to in kWh/year divided by 2000); and MAE was also used for panel count and total energy predictions. 

\section{Additional Qualitative Results}

\begin{figure}
    \centering
    \includegraphics[width=\linewidth]{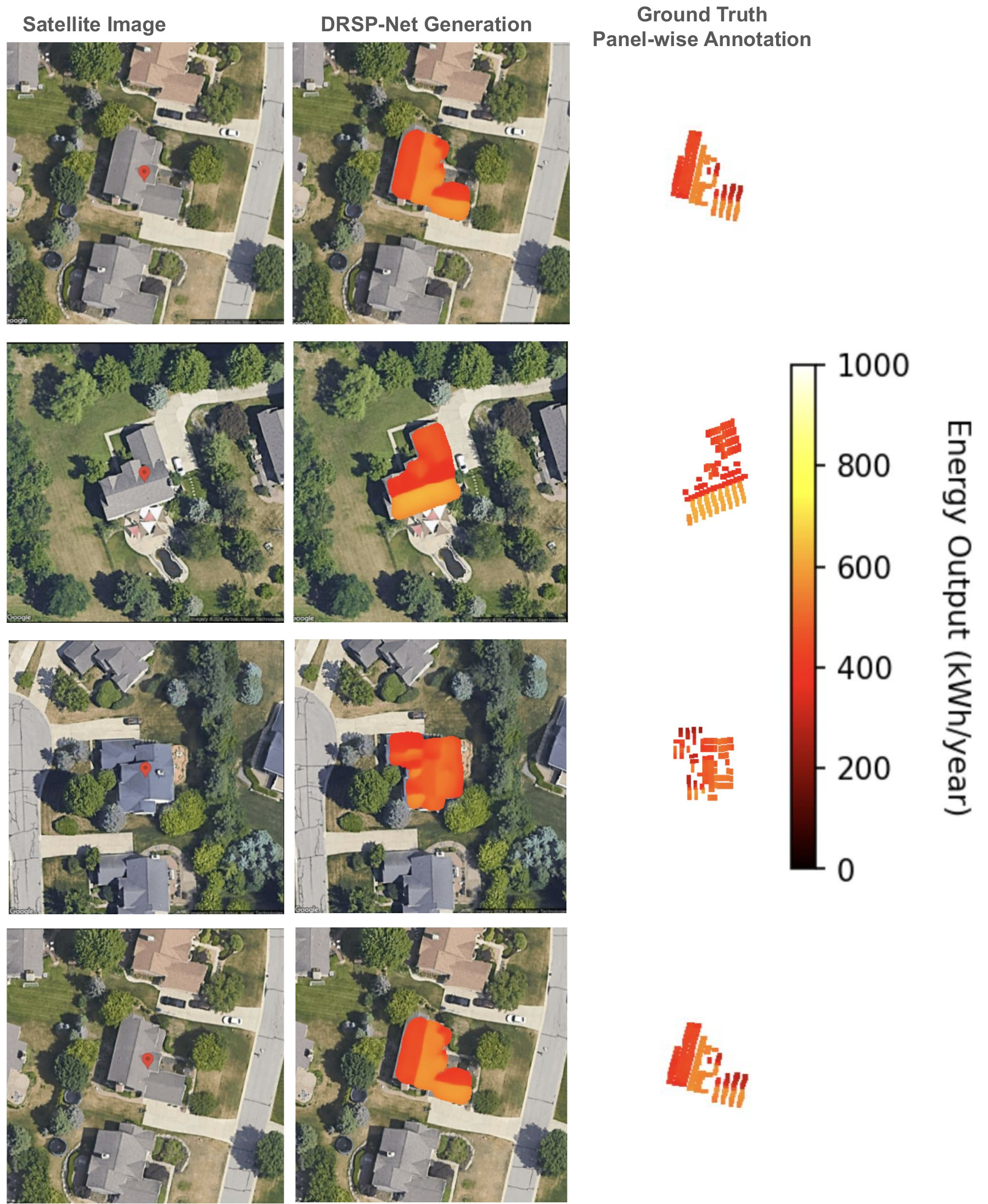}
    \caption{Additional qualitative examples of DRSP-Net (trained on both MIDWEST and TEXAS) generated results. All qualitative examples shown here are from validation and test splits from our dataset.}
    \label{fig:qualapp}
\end{figure}

We show additional qualitative results in Figure~\ref{fig:qualapp}. All qualitative examples shown here are from validation and test splits from our dataset.

\end{document}